\documentclass[10pt,twocolumn,letterpaper]{article}
\usepackage{cvpr} 
\usepackage{multirow}
\usepackage{epsfig}
\usepackage{graphicx}
\usepackage{amsmath}
\usepackage{amssymb}
\usepackage{multirow}
\usepackage{graphics}
\usepackage{threeparttable}
\usepackage{color}
\usepackage[normalem]{ulem}
\usepackage{float}
\usepackage{amsfonts}
\usepackage{bm}
\usepackage{bbm}
\usepackage{enumitem}
\usepackage{algorithmic,algorithm}
\usepackage{subcaption}
\usepackage{array}
\usepackage{colortbl}
\usepackage{pifont}
\usepackage{booktabs}
\usepackage{mathtools}
\usepackage{siunitx}
\usepackage{caption} 
\usepackage[rgb,dvipsnames]{xcolor}
\usepackage[nopar]{lipsum}
\usepackage{listings}
\usepackage[export]{adjustbox}
\usepackage{makecell}
\usepackage[nopar]{lipsum}
\usepackage{listings}

\usepackage{amsmath,amsfonts,bm}

\def\eqref#1{equation~\ref{#1}}

\def\1{\bm{1}}

\def\mX{{\bm{X}}}
\def\mY{{\bm{Y}}}

\DeclareMathAlphabet{\mathsfit}{\encodingdefault}{\sfdefault}{m}{sl}
\SetMathAlphabet{\mathsfit}{bold}{\encodingdefault}{\sfdefault}{bx}{n}

\def\gG{{\mathcal{G}}}

\usepackage{url}
\usepackage{xargs}
\definecolor{cvprblue}{rgb}{0.21,0.49,0.74}
\usepackage[pagebackref,breaklinks,colorlinks,citecolor=cvprblue]{hyperref}

\title{Vision Search Assistant: Empower Vision-Language Models as Multimodal Search Engines}

\author{{Zhixin Zhang}$^{1}$
	\quad {Yiyuan Zhang}$^{1,2\dagger}$
	\quad {Xiaohan Ding}$^{3}$ 
        \quad {Xiangyu Yue}$^{1}$ \\
	\textsuperscript{1} MMLab, CUHK
        \quad 
	\textsuperscript{2} Shanghai AI Laboratory 
        \quad \textsuperscript{3} Tencent\\\\
        \texttt{\url{https://github.com/cnzzx/VSA}}
}

\begin{document}

\maketitle

\begin{abstract}
Search engines enable the retrieval of unknown information with texts. However, traditional methods fall short when it comes to understanding unfamiliar visual content, such as identifying an object that the model has never seen before. This challenge is particularly pronounced for large vision-language models (VLMs): if the model has not been exposed to the object depicted in an image, it struggles to generate reliable answers to the user's question regarding that image. Moreover, as new objects and events continuously emerge, frequently updating VLMs is impractical due to heavy computational burdens. To address this limitation, we propose Vision Search Assistant, a novel framework that facilitates collaboration between VLMs and web agents. This approach leverages VLMs' visual understanding capabilities and web agents' real-time information access to perform open-world Retrieval-Augmented Generation via the web. By integrating visual and textual representations through this collaboration, the model can provide informed responses even when the image is novel to the system. Extensive experiments conducted on both open-set and closed-set QA benchmarks demonstrate that the Vision Search Assistant significantly outperforms the other models and can be widely applied to existing VLMs.
\end{abstract}

\begin{figure*}[ht]
    \centering
    \includegraphics[width=1.0\linewidth]{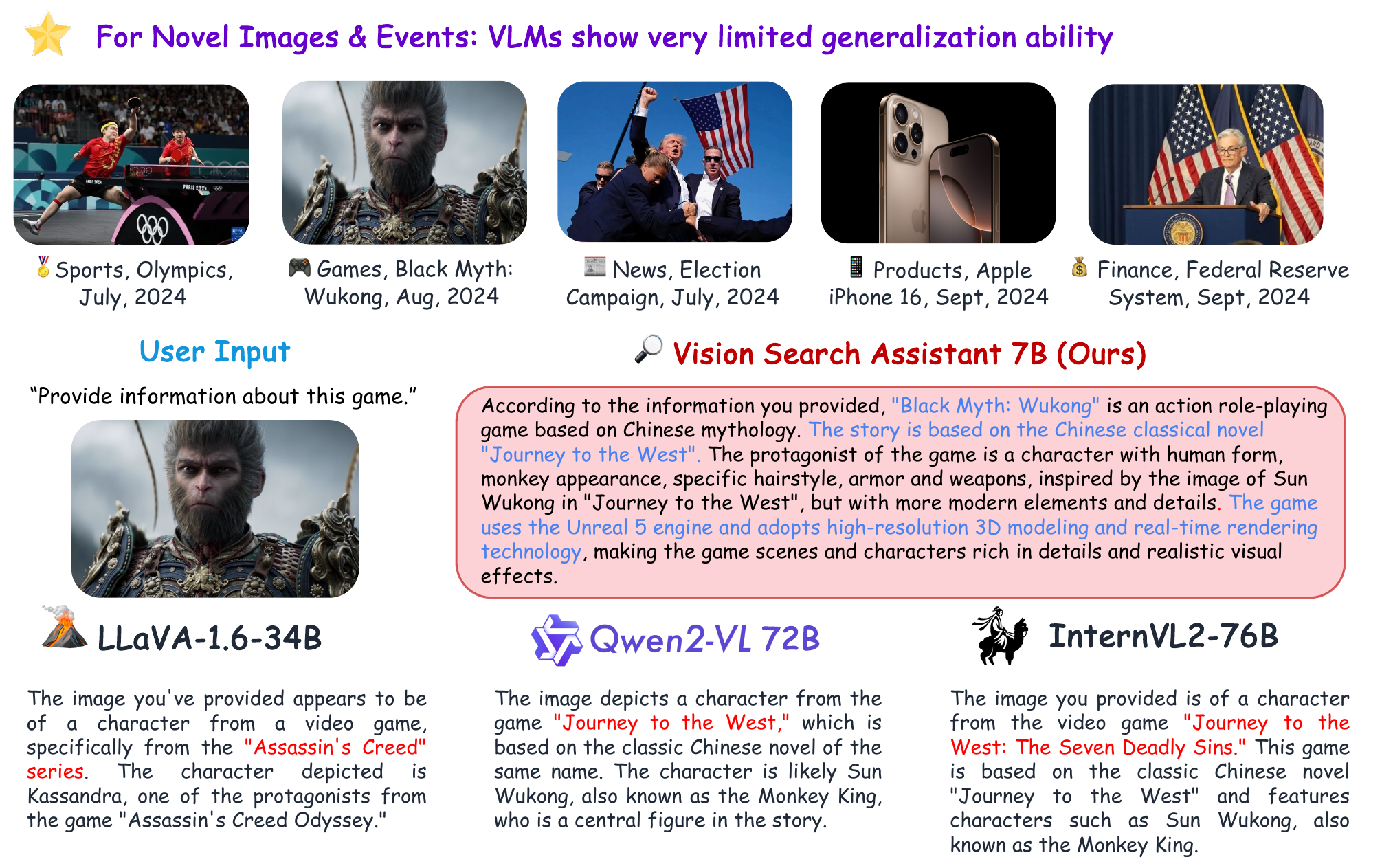}
    \caption{\textbf{Vision Search Assistant acquires unknown visual knowledge through web search}. An intuitive comparison of answering the user's question with an unseen image. The proposed Vision Search Assistant is developed based on LLaVA-1.6-7B, and its ability to answer the question on unseen images outperforms the state-of-the-art models including LLava-1.6-34B~\cite{liu2023improved}, Qwen2-VL-72B~\cite{Qwen-VL}, and InternVL2-76B~\cite{chen2024far}.}
    \label{fig:teaser}
\end{figure*}

\section{Introduction}~\label{sec:intro}
The advent of Large Language Models (LLMs)~\cite{achiam2023gpt,gpt4o,claude3,touvron2023llama,schulman2022chatgpt,vicuna2023} has significantly enhanced the human capacity to acquire unfamiliar knowledge through powerful zero-shot Question-Answering (QA) capabilities. Building upon these advancements, techniques such as Retrieval-Augmented Generation (RAG)~\cite{yu2023improving,shi2023replug,trivedi2022interleaving} have further reinforced LLMs in knowledge-intensive, open-domain QA tasks. Concurrently, recent progress in visual instruction tuning~\cite{liu2023visual,liu2023improved,zhu2023minigpt} has led to the development of large Vision-Language Models (VLMs) that aim to equip LLMs with visual understanding capabilities. By scaling model parameters and training on extensive text-image datasets, VLMs such as LLaVA-1.6-34B~\cite{liu2023improved}, Qwen2-VL-72B~\cite{Qwen-VL}, and InternVL2-76B~\cite{chen2024far} have achieved state-of-the-art performance on the OpenVLM leaderboard\footnote{\url{https://huggingface.co/spaces/opencompass/open_vlm_leaderboard}}. However, LLMs and VLMs are subject to the limitations imposed by their knowledge cutoff dates. They may provide incorrect answers when asked about events or concepts that occurred after their knowledge cutoff dates (Figure~\ref{fig:teaser}) To overcome this limitation for LLMs, they are often connected to web agents~\cite{liu2023webglm,nakano2021webgpt,chen2024agent,deng2024mind2web,bai2024digirl}, which enable internet access and information retrieval, allowing them to obtain the most up-to-date data and improve the accuracy of their responses. Such agents are designed to interpret natural language instructions, navigate complex web environments, and extract relevant textual information from HTML documents, thereby enhancing the accessibility and utility of vast amounts of web-based textual data for a wide range of applications.

\textit{However, for VLMs facing unseen images and novel concepts, their ability to learn and use up-to-date multimodal knowledge from the internet remains a pressing challenge.} As the existing web agents predominantly rely on searching the user's question and summarizing the returned HTML content, they present a notable limitation when handling tasks involving images or other visual content: the visual information is often overlooked or inadequately processed.

In this work, we enable the VLM to answer a question regarding an unseen image or novel concept, which behaves like a human searching the Internet. It \textbf{1)} understands the query, \textbf{2)} decides which objects in the image it should look at and infers the correlations among the objects, \textbf{3)} respectively generates the texts to search, \textbf{4)} analyzes the contents returned from the search engine based on the query and inferred correlations, and \textbf{5)} judges if the obtained visual and textual information is sufficient for generating the answer or it should iterate and refine the above process. Regarding the concrete designs of such a framework, we make contributions by answering the following three questions that remained unanswered in the literature
\begin{itemize}[noitemsep,topsep=0pt,leftmargin=*]
    \item \textit{What to search}: shall we search the descriptions of the whole image or some critical objects?
    \item \textit{How to search}: shall we search once and summarize a huge amount of returned content or search progressively to obtain more related content?
    \item \textit{By what to conclude}: shall the final answer be generated with the eventually summarized web knowledge or all the knowledge acquired through the entire searching process?
\end{itemize}

\begin{figure*}[t]
    \centering
    \includegraphics[width=1.0\linewidth]{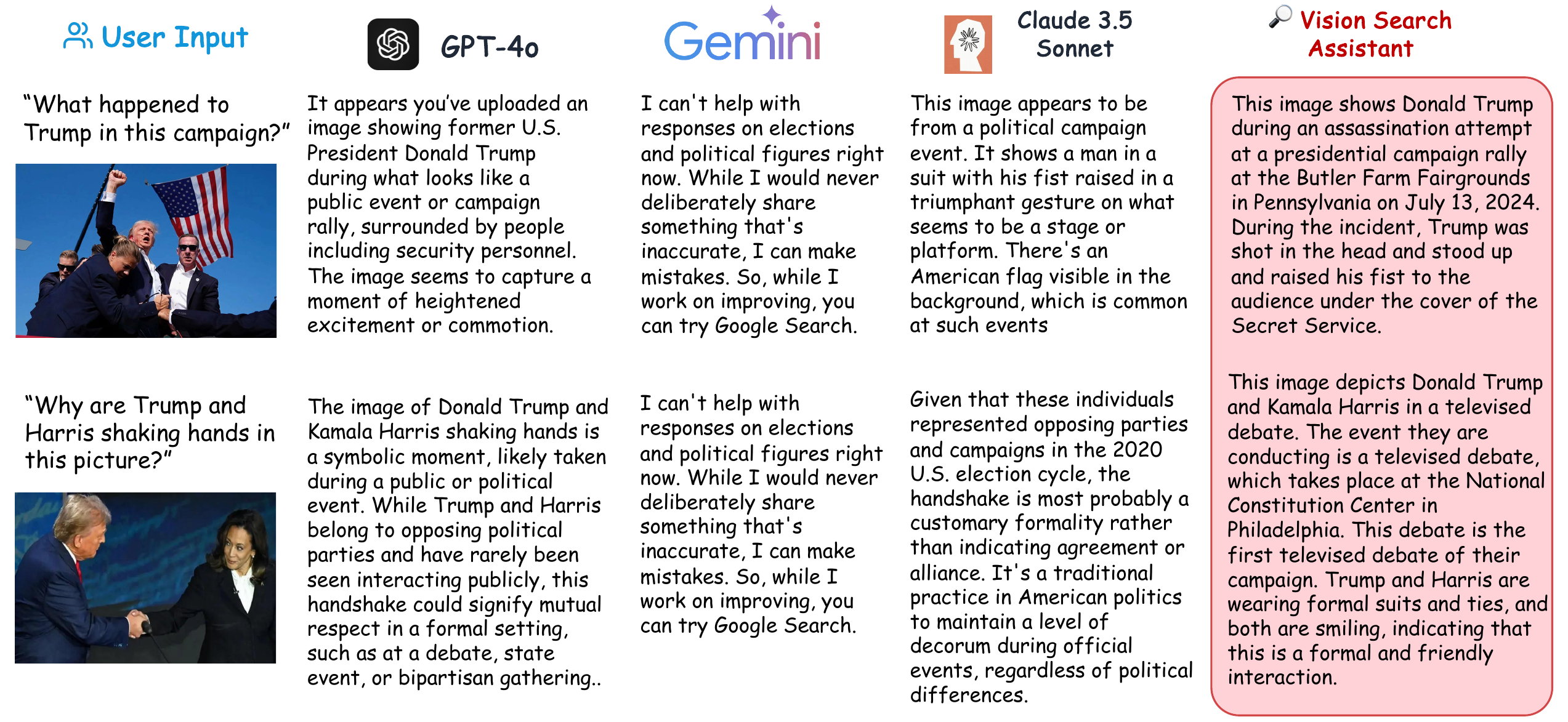}
    \caption{\textbf{Comparsion with Closed-Source Models including GPT-4o~\cite{gpt4o}, Gemini~\cite{reid2024gemini}, Claude 3.5 Sonnet~\cite{claude3} with Vision Search Assistant} shows that Vision Search Assistant satisfies users' needs better even if the image is novel.}
    \label{fig:cmp:close}
    \vspace{-2mm}
\end{figure*}

\begin{figure*}[t]
    \centering
    \includegraphics[width=1.0\linewidth]{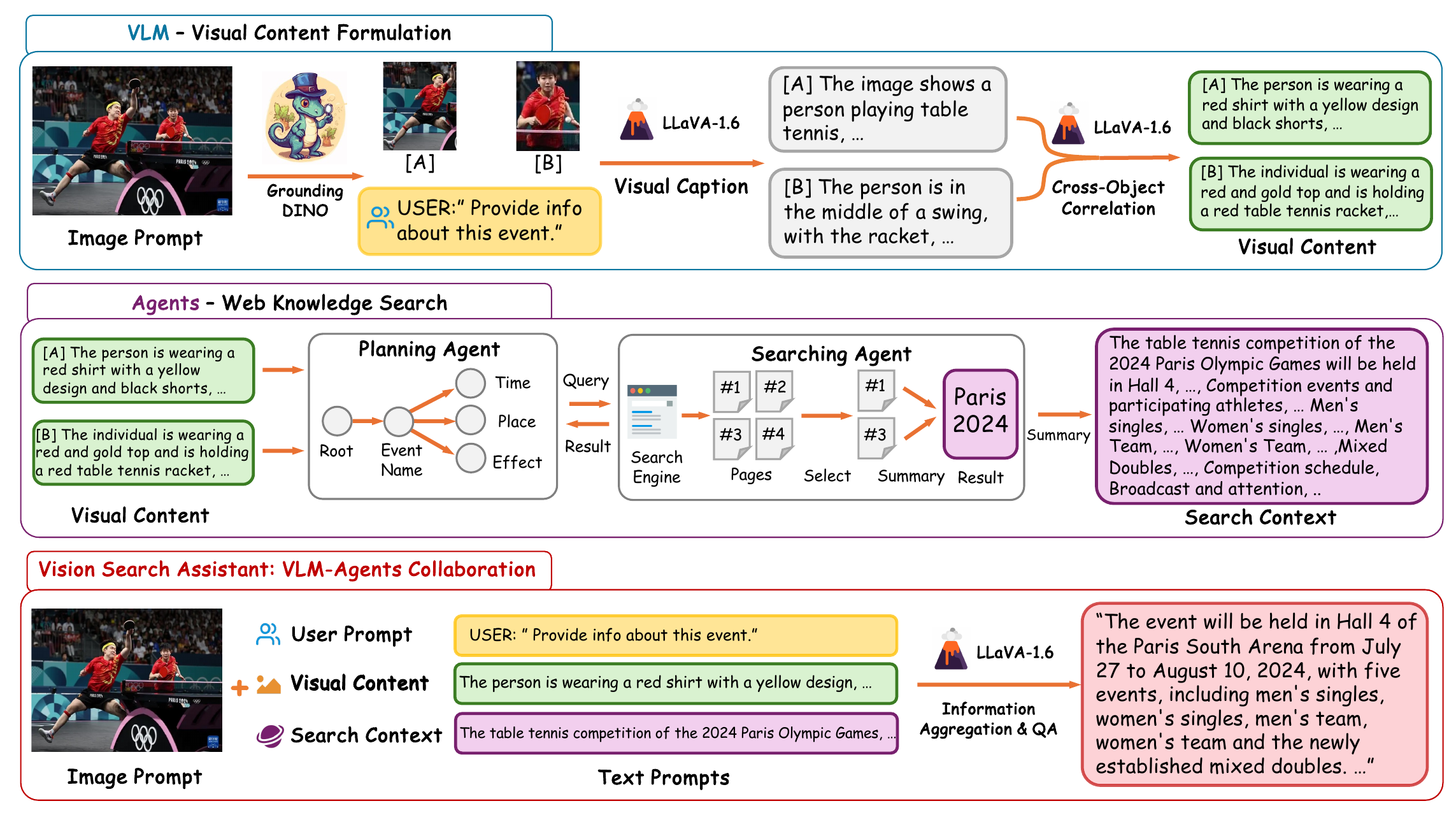}
    \caption{\textbf{Overview of Vision Search Assistant}. We first identify the critical objects and generate their descriptions considering their correlations, named Correlated Formulation, using the Vision Language Model (VLM). We then use the LLM to generate sub-questions that leads to the final answer, which is referred to as the Planning Agent. The web pages returned from the search engine are analyzed, selected, and summarized by the same LLM, which is referred to as the Searching Agent. We use the original image, the user's prompt, the Correlated Formulation together with the obtained web knowledge to generate the final answer. Vision Search Assistant produces reliable answers, even for novel images, by leveraging the collaboration between VLM and web agents to gather visual information from the web effectively.}
    \label{fig:framework}
\end{figure*}

By exploring such three aspects, we propose \textbf{Vision Search Assistant}, a framework based on VLM-agent collaboration, which \textit{empowers an arbitrary VLM to become a multimodal automatic search engine}. We integrate VLMs into web agents to understand what the user wants, where to look, what to search, how to learn from the returned multimodal information, and whether to conclude or search another time. More specifically, Vision Search Assistant conducts three steps (Figure~\ref{fig:framework}): 
\begin{itemize}[noitemsep,topsep=0pt,leftmargin=*]
    \item Visual Content Formulation~(\S~\ref{sec:vsa:vlm}) is proposed to represent the visual content with VLM-generated textural descriptions of critical visual objects and their underlying correlations. Through this step, we obtain a \textit{correlated formulation} for each critical object, which is a textual representation that considers its correlations with other objects.
    \item Web Knowledge Search~(\S~\ref{sec:vsa:agent}) is a novel algorithm that drives the search process. It generates multiple sub-questions with the web agent regarding the user prompt and the correlated formulation of each critical object. Each of such sub-questions can be viewed as a node in a directed graph. For each correlated formulation and each sub-question, we construct the search query by combining the correlated formulation and sub-question and use the LLM to analyze and select useful contents returned from the search engine, then summarize the web knowledge from the answers obtained with all such sub-questions. After that, we iterate the above step by proposing more sub-questions based on the previous sub-questions and known web knowledge, which can be seen as expanding the directed graph. We use the LLM to judge if the latest iteration has obtained sufficient web knowledge to answer the user's question and terminate the process if so. 
    \item Collaborative Generation~(\S~\ref{sec:vsa:co})) is proposed to use the VLM to generate the eventual answer with all the critical objects in the image, the initial question, all of their correlated formulations, and the web knowledge obtained in every iteration.
\end{itemize}

As shown in Figure~\ref{fig:cmp:close}, Vision Search Assistant can generate more precise answers than powerful closed-source models such as GPT-4o~\cite{gpt4o}, Gemini~\cite{reid2024gemini}, and Claude 3.5 Sonnet~\cite{claude3}, which further validates the necessity and promising improvement of VLM-Agents collaboration in tackling the growing complexity of multimodal web data and the rapid influx of novel visual content.
\section{Related Work}~\label{sec:related}
\noindent{\textbf{Vision-Language Models}}.
Pioneering models such as Flamingo~\cite{alayrac2022flamingo}, BLIP-2 \cite{li2023blip}, LLaVA~\cite{liu2023visual}, and MiniGPT-4 \cite{zhu2023minigpt} have been instrumental in training vision-language models for the tasks such as image captioning and visual question answering. Recent works focus on higher-quality datasets~\cite{gong2023multimodal} and developing lightweight, trainable models~\cite{gao2023llama} to enhance efficiency and accessibility. Further progress includes extending large language models (LLMs) to additional modalities and domains, such as audio processing \cite{huang2023language,chen2023x}, and more modalities~\cite{han2024onellm,zhang2023meta,ding2024unireplknet}. Additionally, KOSMOS-2 \cite{peng2023kosmos}, InternVL2~\cite{chen2024far}, MiniGPT-2 \cite{chen2023minigpt}, and LLaVA-1.5~\cite{liu2023improved} incorporate region-level information by encoding visual regions to embeddings of language models. However, despite scaling model parameters and training data, VLMs’ ability to handle unseen images remains limited, as they heavily rely on previously seen text-image pairs. To overcome this, we propose to enhance VLMs’ performance on novel data by improving generalization without relying solely on extensive training pairs.
\noindent{\textbf{Web Search Agents}}.
The development of web search agents has progressed through integrating advanced learning techniques, enhancing autonomy, and optimizing efficiency in web automation. Early models like WebGPT~\cite{nakano2021webgpt} and WebGLM~\cite{liu2023webglm} primarily focused on retrieving information for question-answering tasks, while newer models, such as AutoWebGLM~\cite{lai2024autowebglm}, address deployment challenges with compact designs. Despite their strong web navigation skills, larger models such as WebAgent~\cite{gur2023real} are constrained by size. Incorporating reinforcement learning~\cite{bai2024digirl} and behavior cloning~\cite{zheng2024gpt} has further boosted the efficiency of web agents, as demonstrated by MindAct~\cite{deng2024mind2web}, which integrates cognitive functionalities for complex task execution. While these advances are leading to more scalable and versatile solutions for real-world use, current web agents still struggle with processing visual content directly from the web. We introduce Vision-Language Models to enable web agents to effectively interpret and interact with visual data, significantly expanding their capabilities in handling complex, multimodal tasks. We hope it can make web agents more powerful and adaptable in real-world applications.

\noindent{\textbf{Retrieval-Augmented Generation}}.
Integrating retrieval from large corpora into language models has become essential for knowledge-intensive tasks like open-domain question answering. Instead of relying solely on pre-trained data, the retriever-reader architecture \cite{chen2017reading,guu2020retrieval} enables models to fetch relevant information based on an input query, which the language model then uses to generate accurate predictions. Recent research has enhanced retrievers \cite{karpukhin2020dense,xiong2020approximate,qu2020rocketqa, xiong2020answering,promptrank}, improved readers \cite{izacard2020leveraging,cheng2021unitedqa,yu2021kg, borgeaud2022improving}, jointly fine-tuned both components \cite{yu2022retrieval,Izacard2022FewshotLW,singh2021end, izacard2020distilling}, and integrated retrieval directly within language models \cite{yu2023improving,shi2023replug,trivedi2022interleaving}. 

Therefore, we propose the Vision Search Assistant framework, which introduces an open-world retrieval-augmented generation framework that extends beyond text-based retrieval to operate across both vision and language modalities on the web. It enables VLMs to access real-time, dynamic information, improving their ability to handle novel, cross-modal queries. By pushing the boundaries of retrieval beyond static knowledge sources, we address the challenge of incorporating web-based, multimodal data into generative tasks, offering a more adaptable and scalable solution for RAG.

\section{Vision Search Assistant}~\label{sec:vsa}

\subsection{Visual Content Formulation}~\label{sec:vsa:vlm}
The Visual Content Formulation is proposed to extract the object-level descriptions and correlations among objects in an image. Given the input image \(\mX_{I}\), we first use the open-vocab detector \(\mathcal{F}_{\mathrm{det}}(\cdot)\)~\cite{liu2023grounding} to obtain \(N\) regions of interests in the original image,
\begin{equation}
    \{\mX_{I}^{(i)}\}_{i=1}^{N} = \mathcal{F}_{\mathrm{det}} (\mX_{I}),
\end{equation}
where \(i\) indicates the \(i\)-th region \(\mX_{I}^{(i)}\) in the image \(\mX_{I}\). Then we employ the pretrained VLM~\footnote{Our experiments are conducted with LLaVA-1.6-Vicuna-7B model, which is publicly available at \url{https://huggingface.co/liuhaotian/llava-v1.6-vicuna-7b}.} \(\mathcal{F}_{\mathrm{vlm}}(\cdot,\cdot)\) to caption these regions \(\{\mX_{I}^{(i)}\}_{i=1}^{N}\) conditioned on the tokenized user's textual prompt \(\mX_{T}\), and obtain the visual caption \(\mX_{r}^{(i)}\) for the \(i\)-th region:
\begin{equation}
    \mX_{r}^{(i)} = \mathcal{F}_{\mathrm{vlm}}(\mX_{I}^{(i)},\mX_{T}).
\end{equation}
In this way, we make the regional captions \(\{\mX_{r}^{(i)}\}_{i=1}^N\) contain specific visual information obtained based on the user's interests. To formulate the visual content more comprehensively, we further correlate these visual regions to obtain precise descriptions of the whole image. More specifically, for each region, we concatenate its corresponding caption and the captions of all the other regions. The resultant text, denoted by \([\mX_{r}^{(i)}, \{\mX_{r}^{(j)}\}_{j\ne i}]\), encodes the underlying correlations. It is fed into the VLM together with the image region \(\mX_{I}^{(i)}\). The output is referred to as the \textit{correlated formulation} of each region \(\{\mX_{c}^{(i)}\}_{i=1}^{N}\).
\begin{equation}
    \mX_{c}^{(i)} = \mathcal{F}_{\mathrm{vlm}}(\mX_{I}^{(i)}, [\mX_{r}^{(i)}, \{\mX_{r}^{(j)}\}_{j\ne i}])).
\end{equation}
We will use the correlated formulations of such regions to perform the following web search.

\subsection{Web Knowledge Search: The Chain of Search}~\label{sec:vsa:agent}
The core of Web Knowledge Search is an iterative algorithm named \textit{Chain of Search}, which is designed to obtain the comprehensive web knowledge of the correlated formulations \(\{\mX_{c}^{(i)}\}_{i=1}^{N}\). We take an arbitrary \(i\)-th region \(\mX_{c}^{(i)}\) to elaborate on the Chain of Search algorithm and drop the superscript \((i)\) for convenience.

We use the LLM in our VLM to generate sub-questions that lead to the final answer, which is referred to as the Planning Agent. The web pages returned from the search engine are analyzed, selected, and summarized by the same LLM, which is referred to as the Searching Agent. In this way, we can obtain web knowledge regarding the visual content. Then, based on each of such sub-questions, the Planning Agent generates more sub-questions, and the Searching Agent obtains web knowledge for the next iteration. Formally, we define a directed graph to represent this process, which is \(\gG = \left\langle V, E \right\rangle\), where \(V=\{V_0\}\) is the set of nodes, \(V_0\) is the initial node, and \(E = \varnothing\) is the set of edges. A node represents a set of known information so that \(V_0\) should represent what we know about the region before any web search, \textit{i.e.}, the correlated formulation \(\mX_{c}\). This is formulated as \(V_0 \gets \mX_{c}\). When we search with a sub-question, we will update the graph with a new node representing the web knowledge gained through the sub-question. 

For the \(1\)-st update, we generate sub-questions based on \(V_0\) and denote the generated sub-questions by \((\mX_{s}^{1})=\{(\mX_{s}^{1})_{i}\}_{i=1}^{N_v^1}\), where \(N_v^1\) is the number of sub-questions, \textit{i.e.}, the number of new nodes.

\begin{figure}[ht]
    \centering
    \includegraphics[width=0.98\linewidth]{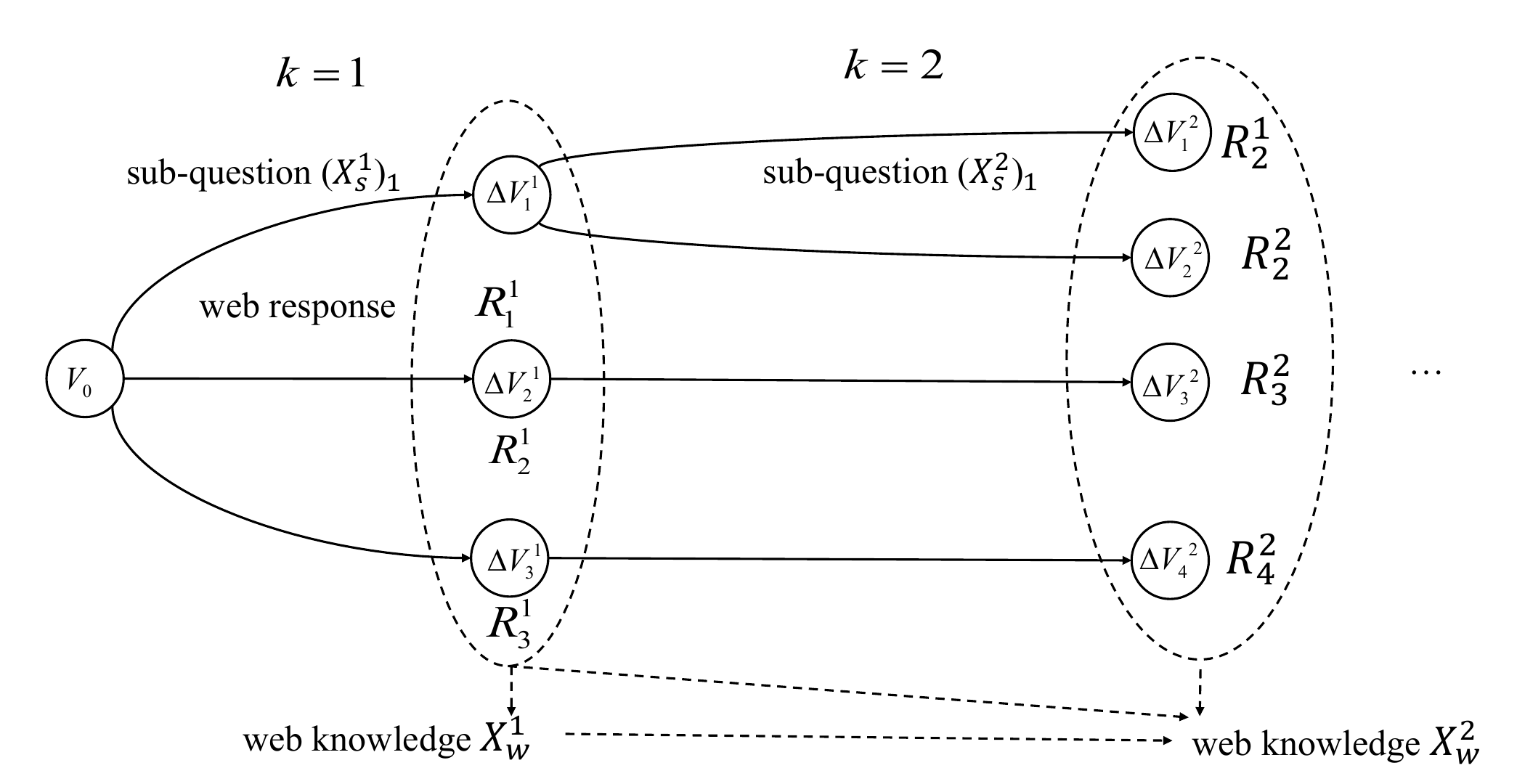}
    \vspace{-2mm}
    \caption{\textbf{The Chain of Search algorithm (\S~\ref{sec:vsa:agent})}. We deduce the update of the directed graph when \(k=1, 2, \cdots\), and the web knowledge is progressively extracted from each update.}
    \vspace{-2mm}
    \label{fig:cos}
\end{figure}

Let \(j\) be the index of the sub-question, the new node \(\Delta V_j^{(1)}\) is a child of \(V_0\), which corresponds to a search with sub-question \((\mX_{s}^{1})_{j}\). The returned set of web pages are formatted as HTML documents. The Searching Agent uses the LLM in our VLM, which is denoted by \(\mathcal{F}_{llm}(\cdot)\), to judge their relevance to the parent node \(V_0\) and the corresponding sub-question \((\mX_{s}^{1})_{j}\) and select those of the highest relevance. The selected web page index \(\tau_j^{1}\) can be formulated
\begin{equation}
\label{eq:search_select_k=1}
    \tau_j^{1} = \mathcal{F}_{\mathrm{llm}}([V_0, (\mX_{s}^{1})_{j}]) \,.
\end{equation}
We use \(\tau_j^{1}\) to select a subset of the HTML documents at the \(1\)-st update, and those selected for sub-question \(j\) are denoted by \(\{P_j^1\}\). We derive the \textit{search response} \(R_j^{1}\) for sub-question \(j\) at the \(1\)-st update by summarizing the selected pages with the LLM, which is \(R_j^{1} = \mathcal{F}_{\mathrm{llm}}(\{P_j^{1}\})\). By the definition of the directed graph, the new node \(\Delta V_j^{(1)}\) should represent \(R_j^{1}\), that is, \(\Delta V_j^{(1)} \gets R_j^{1}\). We add \(\Delta V_j^{(1)}\) into the node set and \((V_0, V_j^{(1)})\) into the edge set. In this paper, \(\Delta V_j^{(1)}\) is synonymous with ``the search response \(R_j^{1}\) obtained with sub-question \((\mX_{s}^{1})_{j}\)''.

\begin{figure*}[t]
    \centering
    \includegraphics[width=0.98\linewidth]{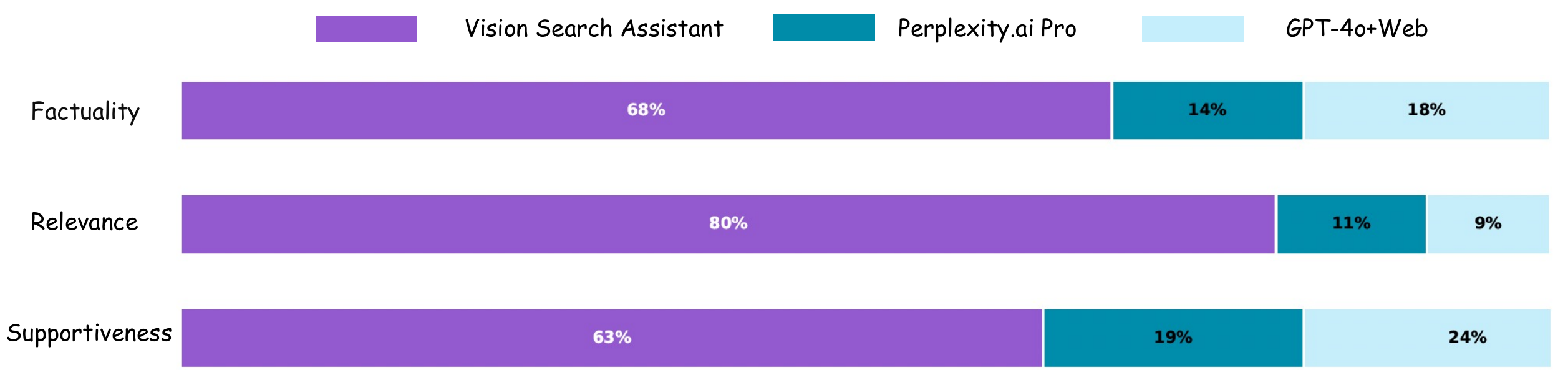}
    \caption{\textbf{Open-Set Evaluation}: We conduct a human expert evaluation on open-set QA tasks. Vision Search Assistant significantly outperformed Perplexity.ai Pro and GPT-4o-Web across three key objectives: factuality, relevance, and supportiveness. }
    \label{tab:open-set-eval}
\end{figure*}

Then, we summarize the search responses of all the \(N_v^1 \) nodes at the \(1\)-st update and obtain the comprehensive \textit{web knowledge} \(\mX_{w}^{(1)}\), which is denoted by 
\begin{equation}
\label{eq:web_know_k=1}
    \mX_{w}^{(1)} = \mathcal{F}_{\mathrm{llm}}([R_1^{1}, R_2^{1}, \cdots, R_{N_v^1}^{1}]) \,.
\end{equation}

\begin{table*}[ht]
    \centering
    \resizebox{0.88\linewidth}{!}{
    \begin{tabular}{lcccc}
    \toprule
     Model & Conversation (\%) & Detail (\%) & Reasoning (\%) & Overall (\%) \\ \hline
     LLava-1.6-7B (Baseline) & 72.9 &76.5 & 84.2 & 78.5 \\
     LLava-1.6-7B (naive search) & 70.3 & 76.7 & 85.8 & 78.9 \\
     LLava-1.6-7B (\textit{w/}~\S~\ref{sec:vsa:agent}) & 72.6 & 78.9 & 89.8 & 82.7 \\
     Vision Search Assistant & \textbf{73.3} (+0.4) & \textbf{79.3} (+2.8) & \textbf{95.0} (+10.8) & \textbf{84.9} (+6.4) \\
     \bottomrule
    \end{tabular}
    }
    \caption{\textbf{Closed-Set Evaluation on the LLaVA-W benchmark}. We use GPT-4o (0806) for evaluation. Naive search here denotes the VLM with Google image search.}
    \label{tab:close-set-eval}
\end{table*}

For the following updates with \(k>1\), we expand the graph similarly but with minor differences:
\begin{itemize}
    \item For each node at update \(k-1\), we use the LLM to generate further sub-questions, just like how we expand \(V_0\) at the \(1\)-st update.
    \item When we select the most relevant web pages for a node \(\Delta V_j^{(k)}\), we analyze their relevance to not only \(V_0\) and the corresponding sub-question \((\mX_{s}^{k})_{j}\) (just like the \(1\)-st update), but also the search response of its parent node.
    \item When we summarize the comprehensive web knowledge \(\mX_{w}^{(k)}\), except for the search responses of all the nodes at the current update, we also use all the known comprehensive web knowledge \(\{ \mX_{w}^{(i)}\}_{i=1}^{k-1}\) and the search responses of all the previous nodes \(\{R_m^n\}_{\{m=1,n=1\}}^{\{m=N_v^n,n=k-1\}}\).
\end{itemize}
Formally, when \(k>1\): 
\begin{equation}
\begin{aligned}
\label{eq:web_know_k>1}
\small
    \tau_j^{k} =& \mathcal{F}_{\mathrm{llm}}([V_0, (\mX_{s}^{k})_{j}, R_i^{k-1}]), \\
    \mX_{w}^{(k)} =& \mathcal{F}_{\mathrm{llm}}(\{ \mX_{w}^{(i)}\}_{i=1}^{k-1}, \{R_m^n\}_{\{m=1,n=1\}}^{\{m=N_v^n,n=k-1\}}, \{R_i^{k}\}_{i=1}^{N_v^k}).
    \end{aligned}
\end{equation}

At each update, the search agent uses the LLM to judge if the knowledge currently obtained is sufficient to answer the initial question. If so, we terminate the process.

\subsection{Collaborative Generation}~\label{sec:vsa:co}
We use the original image $\mX_{I}$, the user's initial prompt $\mX_{T}$, and the Correlated Formulations $\{\mX_{C}^{(i)}\}_{i=1}^{N}$ together with the obtained web knowledge $\{\mX_{W}^{(i)}\}_{i=1}^{N}$ to collaboratively generate the final answer \(\mY\) with the VLM:
\begin{equation}
    \mY = \mathcal{F}_{\mathrm{vlm}}(\mX_{I}, \{\mX_{c}^{(i)}\}_{i=1}^{N}, \{\mX_{w}^{(i)}\}_{i=1}^{N}, \mX_{T}) \,.
\end{equation}
\section{Experiments}~\label{sec:exp}
\vspace{-3mm}
\subsection{Open-Set Evaluation}
\begin{figure*}[t]
    \centering
    \includegraphics[width=0.95\linewidth]{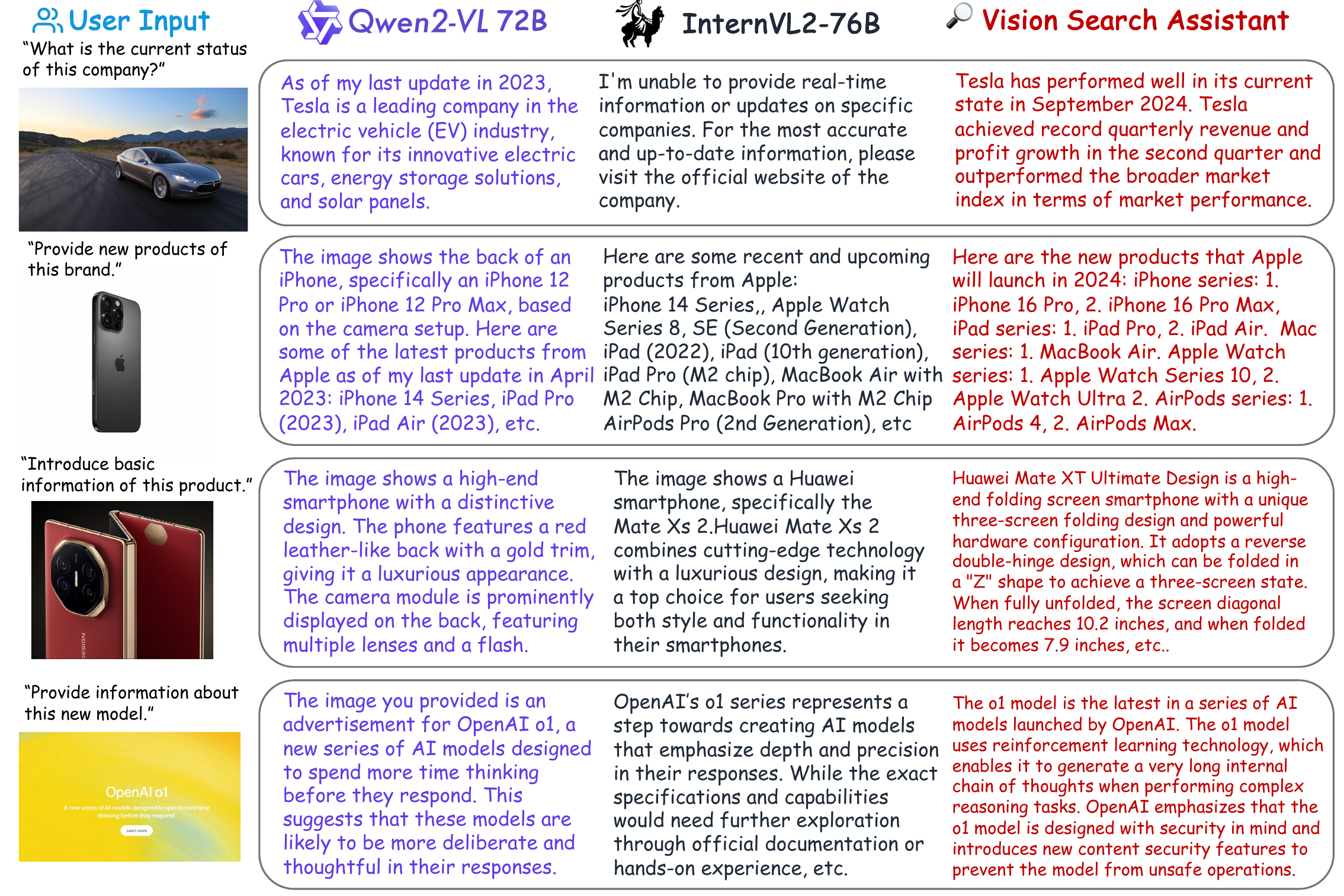}
    \caption{\textbf{Comparisons among Qwen2-VL-72B, InternVL2-76B, and Vision Search Assistant.} We compare the open-set QA results on both novel events (the first two rows) and images (the last two rows). Vision Search Assistant excels in generating accurate and detailed results.}
    \vspace{-3mm}
    \label{fig:demo}
\end{figure*}
\noindent{\textbf{Setup}}. In the Open-Set Evaluation, we performed a comparative assessment by 10 human experts evaluation, which involved questions of 100 image-text pairs collected from the news from July 15th to September 25th covering all fields on both novel images and events. Human experts conducted the evaluations across three critical dimensions: factuality, relevance, and supportiveness.

\begin{figure*}[ht]
    \centering
    \includegraphics[width=0.95\linewidth]{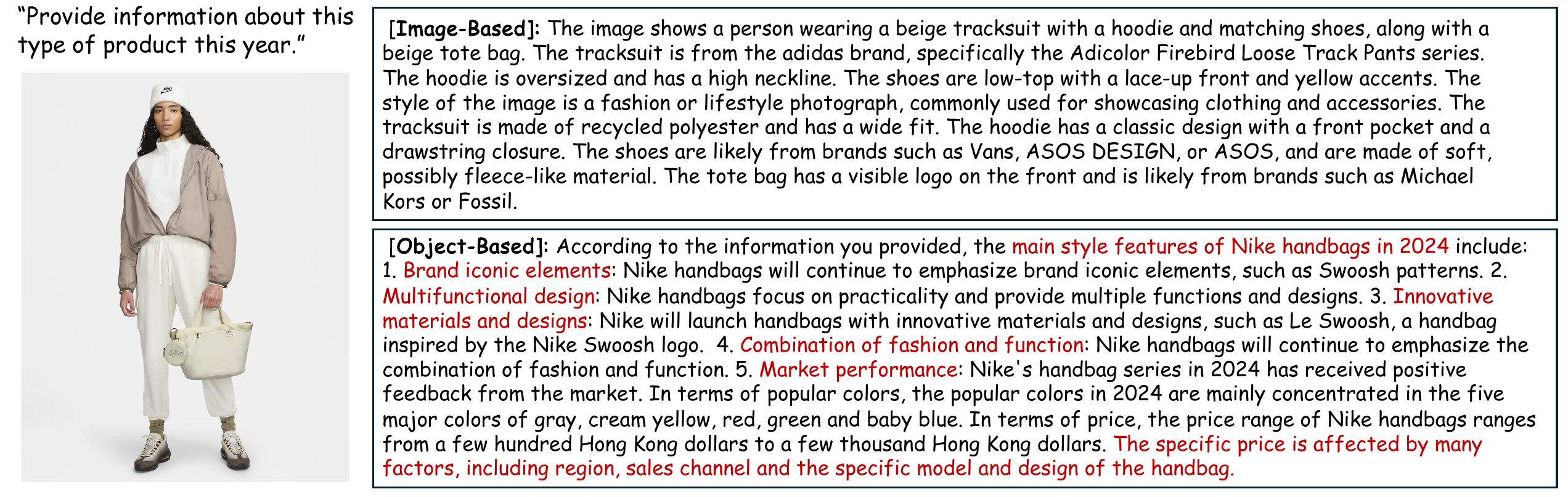}
    \vspace{-2mm}
    \caption{\textbf{Ablation Study on What to Search}. We use the object description to avoid the visual redundancy of the image.}
    \vspace{-4mm}
    \label{fig:ablation-a}
\end{figure*}

\begin{figure*}[ht]
    \centering
    \includegraphics[width=0.98\linewidth]{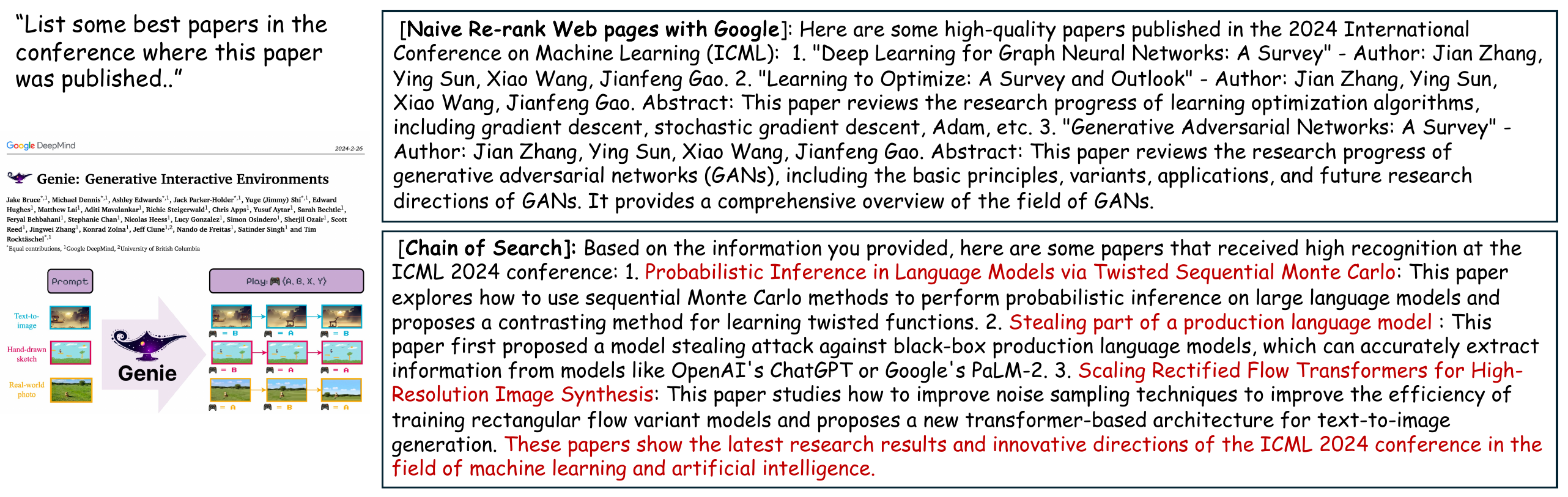}
    \caption{\textbf{Ablation Study on How to search}. We propose the \textit{Chain of Search} (\S~\ref{sec:vsa:agent}) to obtain related web knowledge for VLMs progressively.}
    \vspace{-4mm}
    \label{fig:ablation-b}
\end{figure*}

\begin{figure*}[ht]
    \centering
    \includegraphics[width=0.98\linewidth]{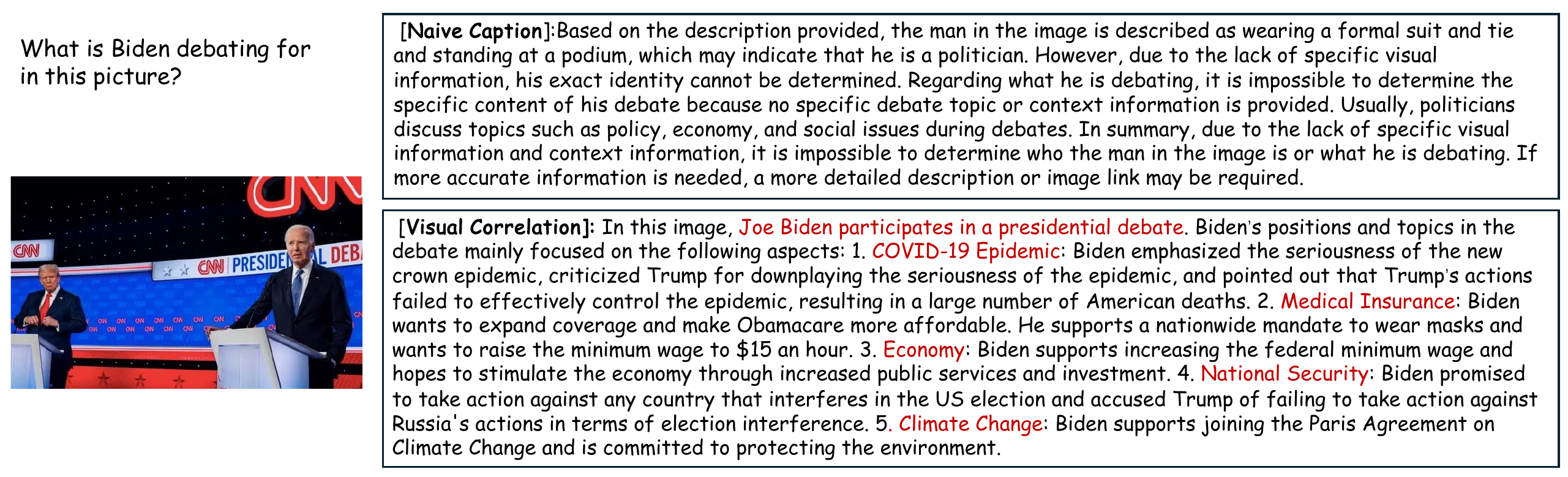}
    \caption{\textbf{Ablation Study on Complex Scenarios}. We use visual correlation to improve the ability in multiple-object scenarios.}
    \label{fig:ablation-c}
    \vspace{-2mm}
\end{figure*}

\begin{figure*}[ht]
    \centering
    \includegraphics[width=0.85\linewidth]{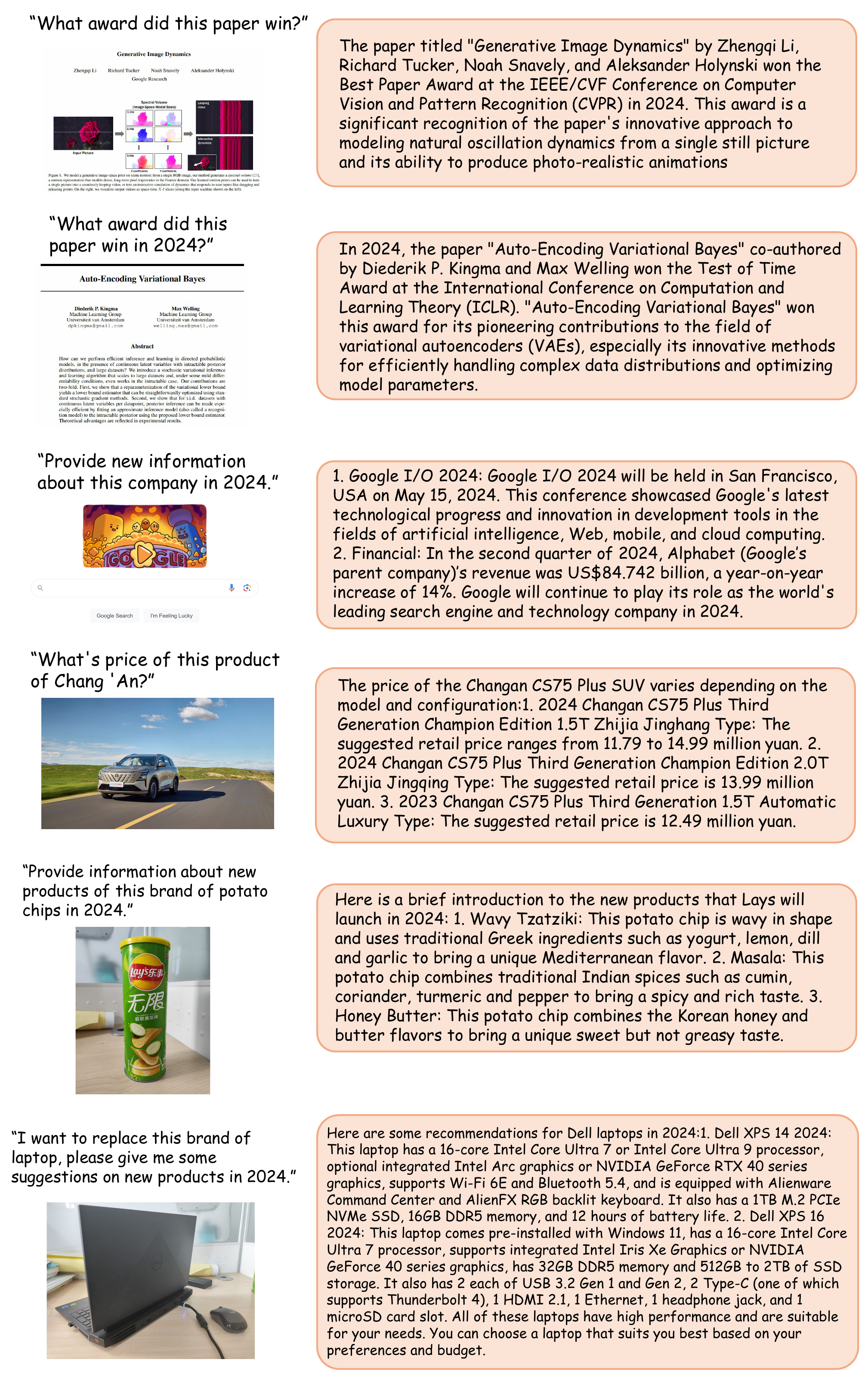}
    \caption{\textbf{A series of demos of Vision Search Assistant on novel images, events, and in-the-wild scenarios.} Vision Search Assistant delivers promising potential as a powerful multimodal engine.}
    \label{fig:longdemo}
\end{figure*}

\noindent{\textbf{Results and Analysis}}. As illustrated in Figure~\ref{tab:open-set-eval}, Vision Search Assistant demonstrated superior performance across all three dimensions compared to Perplexity.ai Pro and GPT-4-Web: \textbf{1}) Factuality: Vision Search Assistant scored 68\%, outperforming Perplexity.ai Pro (14\%) and GPT-4-Web (18\%). This significant lead indicates that Vision Search Assistant consistently provided more accurate and fact-based answers.  \textbf{2}) 	Relevance: With a relevance score of 80\%, Vision Search Assistant demonstrated a substantial advantage in providing highly pertinent answers. In comparison, Perplexity.ai Pro and GPT-4-Web achieved 11\% and 9\%, respectively, showing a significant gap in their ability to maintain topicality with the web search. \textbf{3}) Supportiveness: Vision Search Assistant also outperformed the other models in providing sufficient evidence and justifications for its responses, scoring 63\% in supportiveness. Perplexity.ai Pro and GPT-4-Web trailed with scores of 19\% and 24\%, respectively. These results underscore the superior performance of Vision Search Assistant in open-set tasks, particularly in delivering comprehensive, relevant, and well-supported answers, positioning it as an effective method for handling novel images and events.

\subsection{Closed-Set Evaluation}
\noindent{\textbf{Setup}}. We conduct the closed-set evaluation on the LLaVA-W~\cite{liu2023improved} benchmark, which contains 60 questions regarding the Conversation, Detail, and Reasoning abilities of VLMs in the wild. We use the GPT-4o(0806) model for evaluation. We use LLaVA-1.6-7B as our baseline model, that has been evaluated in two modes: the standard mode and a ``naive search” mode that utilizes a simple Google Image search component. Additionally, an enhanced version of LLaVA-1.6-7B, equipped with improvements outlined in section \S~\ref{sec:vsa:agent}, is also evaluated.

\noindent{\textbf{Results and Analysis}}. As shown in Table~\ref{tab:close-set-eval}, the Vision Search Assistant demonstrates the strongest performance across all categories. Specifically, it achieves a 73.3\% score in the conversation category, representing a modest gain of +0.4\% compared to the LLaVA models. In the detail category, the Vision Search Assistant stands out with a score of 79.3\%, outperforming the highest-performing LLaVA variation by +2.8\%.
When it comes to reasoning, our method brings out +10.8\% above the best-performing LLaVA model. This suggests that the Vision Search Assistant’s advanced integration of visual and textual search greatly enhances its reasoning capabilities. The overall performance of the Vision Search Assistant is 84.9\%, marking an improvement of +6.4\% over the baseline model. This shows that the Vision Search Assistant excels in both conversational and reasoning tasks, giving it a clear advantage for in-the-wild abilities.

\subsection{Ablation Study}
\noindent{\textbf{What to search: Object-Level Descriptions}}. As illustrated in Figure~\ref{fig:ablation-a}, if we use the image-based caption, the search agent can not precisely focus on the key information (the handbag in this figure), meanwhile, the image contains visual redundancy, which obstacles the textual description to drive web agent and retrieve the most relevant web pages, therefore, we use the object-level description in the following ablation study.

\noindent{\textbf{Complex Scenarios of Search: Visual Correlation}}. We find that the caption can not fully support the search ability in multiple-object scenarios. As shown in Figure~\ref{fig:ablation-c}, the caption of Biden can not answer the questions on the group-wise debate, the visual correlation (``debate'' in this demo) between Trump can effectively improve the answer quality.

\noindent{\textbf{How to search}}: \textbf{Chain of Search} (\S~\ref{sec:vsa:agent}). The trivial idea to incorporate web search with VLMs is to introduce a Google search engine and re-rank the large-scale related pages. As shown in Figure~\ref{fig:ablation-b}, we found it difficult to directly obtain the required knowledge since the page-rank method prefers more hyper-link pages instead of exact relevance. The VLM is also limited to its context length to summarize a large number of pages. Therefore, we propose the chain of search and enable the progressive summary of web knowledge aiming to answer the user's questions.

\section{Conclusion and Discussion}~\label{sec:conclusion}
In this paper, we seek to improve the generalization ability of VLMs of novel images and extend the capacity of web agents to solve visual content tasks. Through the synergistic collaboration between VLMs and web agents, we find that VLMs can generate more reliable answers regarding novel images with the help of real-time web knowledge retrieval, and web agents can solve more challenging tasks than HTML documents only. Meanwhile, there are also some limitations inside the Vision Search Assistant framework such as the exact inference speed of VLMs, the web condition of web agents, and the retrieval efficiency. We hope this paper can inspire more research to address the challenges of VLMs in user experience and improve the automation abilities of web agents across diverse modalities.

{
    \small
    \bibliographystyle{ieeenat_fullname}
    \bibliography{egbib}

\begin{thebibliography}{50}
\providecommand{\natexlab}[1]{#1}
\providecommand{\url}[1]{\texttt{#1}}
\expandafter\ifx\csname urlstyle\endcsname\relax
  \providecommand{\doi}[1]{doi: #1}\else
  \providecommand{\doi}{doi: \begingroup \urlstyle{rm}\Url}\fi

\bibitem[Achiam et~al.(2023)Achiam, Adler, Agarwal, Ahmad, Akkaya, Aleman, Almeida, Altenschmidt, Altman, Anadkat, et~al.]{achiam2023gpt}
Josh Achiam, Steven Adler, Sandhini Agarwal, Lama Ahmad, Ilge Akkaya, Florencia~Leoni Aleman, Diogo Almeida, Janko Altenschmidt, Sam Altman, Shyamal Anadkat, et~al.
\newblock Gpt-4 technical report.
\newblock \emph{arXiv preprint arXiv:2303.08774}, 2023.

\bibitem[Alayrac et~al.(2022)Alayrac, Donahue, Luc, Miech, Barr, Hasson, Lenc, Mensch, Millican, Reynolds, et~al.]{alayrac2022flamingo}
Jean-Baptiste Alayrac, Jeff Donahue, Pauline Luc, Antoine Miech, Iain Barr, Yana Hasson, Karel Lenc, Arthur Mensch, Katherine Millican, Malcolm Reynolds, et~al.
\newblock Flamingo: a visual language model for few-shot learning.
\newblock \emph{Advances in Neural Information Processing Systems}, 35:\penalty0 23716--23736, 2022.

\bibitem[Anthropic(2024)]{claude3}
Anthropic.
\newblock Introducing the next generation of claude.
\newblock 2024.

\bibitem[Bai et~al.(2024)Bai, Zhou, Cemri, Pan, Suhr, Levine, and Kumar]{bai2024digirl}
Hao Bai, Yifei Zhou, Mert Cemri, Jiayi Pan, Alane Suhr, Sergey Levine, and Aviral Kumar.
\newblock Digirl: Training in-the-wild device-control agents with autonomous reinforcement learning.
\newblock \emph{arXiv preprint arXiv:2406.11896}, 2024.

\bibitem[Bai et~al.(2023)Bai, Bai, Yang, Wang, Tan, Wang, Lin, Zhou, and Zhou]{Qwen-VL}
Jinze Bai, Shuai Bai, Shusheng Yang, Shijie Wang, Sinan Tan, Peng Wang, Junyang Lin, Chang Zhou, and Jingren Zhou.
\newblock Qwen-vl: A versatile vision-language model for understanding, localization, text reading, and beyond.
\newblock \emph{arXiv preprint arXiv:2308.12966}, 2023.

\bibitem[Borgeaud et~al.(2022)Borgeaud, Mensch, Hoffmann, Cai, Rutherford, Millican, Van Den~Driessche, Lespiau, Damoc, Clark, et~al.]{borgeaud2022improving}
Sebastian Borgeaud, Arthur Mensch, Jordan Hoffmann, Trevor Cai, Eliza Rutherford, Katie Millican, George~Bm Van Den~Driessche, Jean-Baptiste Lespiau, Bogdan Damoc, Aidan Clark, et~al.
\newblock Improving language models by retrieving from trillions of tokens.
\newblock In \emph{International conference on machine learning}, pages 2206--2240. PMLR, 2022.

\bibitem[Chen et~al.(2017)Chen, Fisch, Weston, and Bordes]{chen2017reading}
Danqi Chen, Adam Fisch, Jason Weston, and Antoine Bordes.
\newblock Reading wikipedia to answer open-domain questions.
\newblock In \emph{Proceedings of the 55th Annual Meeting of the Association for Computational Linguistics (Volume 1: Long Papers)}, pages 1870--1879, 2017.

\bibitem[Chen et~al.(2023{\natexlab{a}})Chen, Han, Zhao, Zhang, Shi, Xu, and Xu]{chen2023x}
Feilong Chen, Minglun Han, Haozhi Zhao, Qingyang Zhang, Jing Shi, Shuang Xu, and Bo Xu.
\newblock X-llm: Bootstrapping advanced large language models by treating multi-modalities as foreign languages.
\newblock \emph{arXiv preprint arXiv:2305.04160}, 2023{\natexlab{a}}.

\bibitem[Chen et~al.(2023{\natexlab{b}})Chen, Zhu, Shen, Li, Liu, Zhang, Krishnamoorthi, Chandra, Xiong, and Elhoseiny]{chen2023minigpt}
Jun Chen, Deyao Zhu, Xiaoqian Shen, Xiang Li, Zechun Liu, Pengchuan Zhang, Raghuraman Krishnamoorthi, Vikas Chandra, Yunyang Xiong, and Mohamed Elhoseiny.
\newblock Minigpt-v2: large language model as a unified interface for vision-language multi-task learning.
\newblock \emph{arXiv preprint arXiv:2310.09478}, 2023{\natexlab{b}}.

\bibitem[Chen et~al.(2024{\natexlab{a}})Chen, Liu, Wang, Zhang, Liu, Lin, Chen, and Zhao]{chen2024agent}
Zehui Chen, Kuikun Liu, Qiuchen Wang, Wenwei Zhang, Jiangning Liu, Dahua Lin, Kai Chen, and Feng Zhao.
\newblock Agent-flan: Designing data and methods of effective agent tuning for large language models.
\newblock \emph{arXiv preprint arXiv:2403.12881}, 2024{\natexlab{a}}.

\bibitem[Chen et~al.(2024{\natexlab{b}})Chen, Wang, Tian, Ye, Gao, Cui, Tong, Hu, Luo, Ma, et~al.]{chen2024far}
Zhe Chen, Weiyun Wang, Hao Tian, Shenglong Ye, Zhangwei Gao, Erfei Cui, Wenwen Tong, Kongzhi Hu, Jiapeng Luo, Zheng Ma, et~al.
\newblock How far are we to gpt-4v? closing the gap to commercial multimodal models with open-source suites.
\newblock \emph{arXiv preprint arXiv:2404.16821}, 2024{\natexlab{b}}.

\bibitem[Cheng et~al.(2021)Cheng, Shen, Liu, He, Chen, and Gao]{cheng2021unitedqa}
Hao Cheng, Yelong Shen, Xiaodong Liu, Pengcheng He, Weizhu Chen, and Jianfeng Gao.
\newblock Unitedqa: A hybrid approach for open domain question answering.
\newblock \emph{arXiv preprint arXiv:2101.00178}, 2021.

\bibitem[Chiang et~al.(2023)Chiang, Li, Lin, Sheng, Wu, Zhang, Zheng, Zhuang, Zhuang, Gonzalez, Stoica, and Xing]{vicuna2023}
Wei-Lin Chiang, Zhuohan Li, Zi Lin, Ying Sheng, Zhanghao Wu, Hao Zhang, Lianmin Zheng, Siyuan Zhuang, Yonghao Zhuang, Joseph~E. Gonzalez, Ion Stoica, and Eric~P. Xing.
\newblock Vicuna: An open-source chatbot impressing gpt-4 with 90\%* chatgpt quality, 2023.

\bibitem[Deng et~al.(2024)Deng, Gu, Zheng, Chen, Stevens, Wang, Sun, and Su]{deng2024mind2web}
Xiang Deng, Yu Gu, Boyuan Zheng, Shijie Chen, Sam Stevens, Boshi Wang, Huan Sun, and Yu Su.
\newblock Mind2web: Towards a generalist agent for the web.
\newblock \emph{Advances in Neural Information Processing Systems}, 36, 2024.

\bibitem[Ding et~al.(2024)Ding, Zhang, Ge, Zhao, Song, Yue, and Shan]{ding2024unireplknet}
Xiaohan Ding, Yiyuan Zhang, Yixiao Ge, Sijie Zhao, Lin Song, Xiangyu Yue, and Ying Shan.
\newblock Unireplknet: A universal perception large-kernel convnet for audio video point cloud time-series and image recognition.
\newblock In \emph{Proceedings of the IEEE/CVF Conference on Computer Vision and Pattern Recognition}, pages 5513--5524, 2024.

\bibitem[Gao et~al.(2023)Gao, Han, Zhang, Lin, Geng, Zhou, Zhang, Lu, He, Yue, et~al.]{gao2023llama}
Peng Gao, Jiaming Han, Renrui Zhang, Ziyi Lin, Shijie Geng, Aojun Zhou, Wei Zhang, Pan Lu, Conghui He, Xiangyu Yue, et~al.
\newblock Llama-adapter v2: Parameter-efficient visual instruction model.
\newblock \emph{arXiv preprint arXiv:2304.15010}, 2023.

\bibitem[Gong et~al.(2023)Gong, Lyu, Zhang, Wang, Zheng, Zhao, Liu, Zhang, Luo, and Chen]{gong2023multimodal}
Tao Gong, Chengqi Lyu, Shilong Zhang, Yudong Wang, Miao Zheng, Qian Zhao, Kuikun Liu, Wenwei Zhang, Ping Luo, and Kai Chen.
\newblock Multimodal-gpt: A vision and language model for dialogue with humans.
\newblock \emph{arXiv preprint arXiv:2305.04790}, 2023.

\bibitem[Gur et~al.(2023)Gur, Furuta, Huang, Safdari, Matsuo, Eck, and Faust]{gur2023real}
Izzeddin Gur, Hiroki Furuta, Austin Huang, Mustafa Safdari, Yutaka Matsuo, Douglas Eck, and Aleksandra Faust.
\newblock A real-world webagent with planning, long context understanding, and program synthesis.
\newblock \emph{arXiv preprint arXiv:2307.12856}, 2023.

\bibitem[Guu et~al.(2020)Guu, Lee, Tung, Pasupat, and Chang]{guu2020retrieval}
Kelvin Guu, Kenton Lee, Zora Tung, Panupong Pasupat, and Mingwei Chang.
\newblock Retrieval augmented language model pre-training.
\newblock In \emph{International conference on machine learning}, pages 3929--3938. PMLR, 2020.

\bibitem[Han et~al.(2024)Han, Gong, Zhang, Wang, Zhang, Lin, Qiao, Gao, and Yue]{han2024onellm}
Jiaming Han, Kaixiong Gong, Yiyuan Zhang, Jiaqi Wang, Kaipeng Zhang, Dahua Lin, Yu Qiao, Peng Gao, and Xiangyu Yue.
\newblock Onellm: One framework to align all modalities with language.
\newblock In \emph{Proceedings of the IEEE/CVF Conference on Computer Vision and Pattern Recognition}, pages 26584--26595, 2024.

\bibitem[Huang et~al.(2023)Huang, Dong, Wang, Hao, Singhal, Ma, Lv, Cui, Mohammed, Liu, et~al.]{huang2023language}
Shaohan Huang, Li Dong, Wenhui Wang, Yaru Hao, Saksham Singhal, Shuming Ma, Tengchao Lv, Lei Cui, Owais~Khan Mohammed, Qiang Liu, et~al.
\newblock Language is not all you need: Aligning perception with language models.
\newblock \emph{arXiv preprint arXiv:2302.14045}, 2023.

\bibitem[Izacard and Grave(2020{\natexlab{a}})]{izacard2020distilling}
Gautier Izacard and Edouard Grave.
\newblock Distilling knowledge from reader to retriever for question answering.
\newblock \emph{arXiv preprint arXiv:2012.04584}, 2020{\natexlab{a}}.

\bibitem[Izacard and Grave(2020{\natexlab{b}})]{izacard2020leveraging}
Gautier Izacard and Edouard Grave.
\newblock Leveraging passage retrieval with generative models for open domain question answering.
\newblock \emph{arXiv preprint arXiv:2007.01282}, 2020{\natexlab{b}}.

\bibitem[Izacard et~al.(2022)Izacard, Lewis, Lomeli, Hosseini, Petroni, Schick, Yu, Joulin, Riedel, and Grave]{Izacard2022FewshotLW}
Gautier Izacard, Patrick Lewis, Maria Lomeli, Lucas Hosseini, Fabio Petroni, Timo Schick, Jane~A. Yu, Armand Joulin, Sebastian Riedel, and Edouard Grave.
\newblock Few-shot learning with retrieval augmented language models.
\newblock \emph{ArXiv}, abs/2208.03299, 2022.

\bibitem[Karpukhin et~al.(2020)Karpukhin, O{\u{g}}uz, Min, Lewis, Wu, Edunov, Chen, and Yih]{karpukhin2020dense}
Vladimir Karpukhin, Barlas O{\u{g}}uz, Sewon Min, Patrick Lewis, Ledell Wu, Sergey Edunov, Danqi Chen, and Wen-tau Yih.
\newblock Dense passage retrieval for open-domain question answering.
\newblock \emph{arXiv preprint arXiv:2004.04906}, 2020.

\bibitem[Khalifa et~al.(2023)Khalifa, Logeswaran, Lee, Lee, and Wang]{promptrank}
Muhammad Khalifa, Lajanugen Logeswaran, Moontae Lee, Honglak Lee, and Lu Wang.
\newblock Few-shot reranking for multi-hop qa via language model prompting.
\newblock \emph{arXiv preprint arXiv:2205.12650}, 2023.

\bibitem[Lai et~al.(2024)Lai, Liu, Iong, Yao, Chen, Shen, Yu, Zhang, Zhang, Dong, et~al.]{lai2024autowebglm}
Hanyu Lai, Xiao Liu, Iat~Long Iong, Shuntian Yao, Yuxuan Chen, Pengbo Shen, Hao Yu, Hanchen Zhang, Xiaohan Zhang, Yuxiao Dong, et~al.
\newblock Autowebglm: Bootstrap and reinforce a large language model-based web navigating agent.
\newblock \emph{arXiv preprint arXiv:2404.03648}, 2024.

\bibitem[Li et~al.(2023)Li, Li, Savarese, and Hoi]{li2023blip}
Junnan Li, Dongxu Li, Silvio Savarese, and Steven Hoi.
\newblock Blip-2: Bootstrapping language-image pre-training with frozen image encoders and large language models.
\newblock \emph{arXiv preprint arXiv:2301.12597}, 2023.

\bibitem[Liu et~al.(2023{\natexlab{a}})Liu, Li, Li, and Lee]{liu2023improved}
Haotian Liu, Chunyuan Li, Yuheng Li, and Yong~Jae Lee.
\newblock Improved baselines with visual instruction tuning.
\newblock \emph{arXiv preprint arXiv:2310.03744}, 2023{\natexlab{a}}.

\bibitem[Liu et~al.(2023{\natexlab{b}})Liu, Li, Wu, and Lee]{liu2023visual}
Haotian Liu, Chunyuan Li, Qingyang Wu, and Yong~Jae Lee.
\newblock Visual instruction tuning.
\newblock \emph{arXiv preprint arXiv:2304.08485}, 2023{\natexlab{b}}.

\bibitem[Liu et~al.(2023{\natexlab{c}})Liu, Zeng, Ren, Li, Zhang, Yang, Li, Yang, Su, Zhu, et~al.]{liu2023grounding}
Shilong Liu, Zhaoyang Zeng, Tianhe Ren, Feng Li, Hao Zhang, Jie Yang, Chunyuan Li, Jianwei Yang, Hang Su, Jun Zhu, et~al.
\newblock Grounding dino: Marrying dino with grounded pre-training for open-set object detection.
\newblock \emph{arXiv preprint arXiv:2303.05499}, 2023{\natexlab{c}}.

\bibitem[Liu et~al.(2023{\natexlab{d}})Liu, Lai, Yu, Xu, Zeng, Du, Zhang, Dong, and Tang]{liu2023webglm}
Xiao Liu, Hanyu Lai, Hao Yu, Yifan Xu, Aohan Zeng, Zhengxiao Du, Peng Zhang, Yuxiao Dong, and Jie Tang.
\newblock Webglm: Towards an efficient web-enhanced question answering system with human preferences.
\newblock In \emph{Proceedings of the 29th ACM SIGKDD Conference on Knowledge Discovery and Data Mining}, pages 4549--4560, 2023{\natexlab{d}}.

\bibitem[Nakano et~al.(2021)Nakano, Hilton, Balaji, Wu, Ouyang, Kim, Hesse, Jain, Kosaraju, Saunders, et~al.]{nakano2021webgpt}
Reiichiro Nakano, Jacob Hilton, Suchir Balaji, Jeff Wu, Long Ouyang, Christina Kim, Christopher Hesse, Shantanu Jain, Vineet Kosaraju, William Saunders, et~al.
\newblock Webgpt: Browser-assisted question-answering with human feedback.
\newblock \emph{arXiv preprint arXiv:2112.09332}, 2021.

\bibitem[OpenAI(2024)]{gpt4o}
OpenAI.
\newblock Hello gpt4-o.
\newblock 2024.

\bibitem[Peng et~al.(2023)Peng, Wang, Dong, Hao, Huang, Ma, and Wei]{peng2023kosmos}
Zhiliang Peng, Wenhui Wang, Li Dong, Yaru Hao, Shaohan Huang, Shuming Ma, and Furu Wei.
\newblock Kosmos-2: Grounding multimodal large language models to the world.
\newblock \emph{arXiv preprint arXiv:2306.14824}, 2023.

\bibitem[Qu et~al.(2020)Qu, Ding, Liu, Liu, Ren, Zhao, Dong, Wu, and Wang]{qu2020rocketqa}
Yingqi Qu, Yuchen Ding, Jing Liu, Kai Liu, Ruiyang Ren, Wayne~Xin Zhao, Daxiang Dong, Hua Wu, and Haifeng Wang.
\newblock Rocketqa: An optimized training approach to dense passage retrieval for open-domain question answering.
\newblock \emph{arXiv preprint arXiv:2010.08191}, 2020.

\bibitem[Reid et~al.(2024)Reid, Savinov, Teplyashin, Lepikhin, Lillicrap, Alayrac, Soricut, Lazaridou, Firat, Schrittwieser, et~al.]{reid2024gemini}
Machel Reid, Nikolay Savinov, Denis Teplyashin, Dmitry Lepikhin, Timothy Lillicrap, Jean-baptiste Alayrac, Radu Soricut, Angeliki Lazaridou, Orhan Firat, Julian Schrittwieser, et~al.
\newblock Gemini 1.5: Unlocking multimodal understanding across millions of tokens of context.
\newblock \emph{arXiv preprint arXiv:2403.05530}, 2024.

\bibitem[Schulman et~al.(2022)Schulman, Zoph, Kim, Hilton, Menick, Weng, Uribe, Fedus, Metz, Pokorny, et~al.]{schulman2022chatgpt}
John Schulman, Barret Zoph, Christina Kim, Jacob Hilton, Jacob Menick, Jiayi Weng, Juan Felipe~Ceron Uribe, Liam Fedus, Luke Metz, Michael Pokorny, et~al.
\newblock Chatgpt: Optimizing language models for dialogue.
\newblock \emph{OpenAI blog}, 2022.

\bibitem[Shi et~al.(2023)Shi, Min, Yasunaga, Seo, James, Lewis, Zettlemoyer, and Yih]{shi2023replug}
Weijia Shi, Sewon Min, Michihiro Yasunaga, Minjoon Seo, Rich James, Mike Lewis, Luke Zettlemoyer, and Wen-tau Yih.
\newblock Replug: Retrieval-augmented black-box language models.
\newblock \emph{arXiv preprint arXiv:2301.12652}, 2023.

\bibitem[Singh et~al.(2021)Singh, Reddy, Hamilton, Dyer, and Yogatama]{singh2021end}
Devendra Singh, Siva Reddy, Will Hamilton, Chris Dyer, and Dani Yogatama.
\newblock End-to-end training of multi-document reader and retriever for open-domain question answering.
\newblock \emph{Advances in Neural Information Processing Systems}, 34:\penalty0 25968--25981, 2021.

\bibitem[Touvron et~al.(2023)Touvron, Martin, Stone, Albert, Almahairi, Babaei, Bashlykov, Batra, Bhargava, Bhosale, et~al.]{touvron2023llama}
Hugo Touvron, Louis Martin, Kevin Stone, Peter Albert, Amjad Almahairi, Yasmine Babaei, Nikolay Bashlykov, Soumya Batra, Prajjwal Bhargava, Shruti Bhosale, et~al.
\newblock Llama 2: Open foundation and fine-tuned chat models.
\newblock \emph{arXiv preprint arXiv:2307.09288}, 2023.

\bibitem[Trivedi et~al.(2022)Trivedi, Balasubramanian, Khot, and Sabharwal]{trivedi2022interleaving}
Harsh Trivedi, Niranjan Balasubramanian, Tushar Khot, and Ashish Sabharwal.
\newblock Interleaving retrieval with chain-of-thought reasoning for knowledge-intensive multi-step questions.
\newblock \emph{arXiv preprint arXiv:2212.10509}, 2022.

\bibitem[Xiong et~al.(2020{\natexlab{a}})Xiong, Xiong, Li, Tang, Liu, Bennett, Ahmed, and Overwijk]{xiong2020approximate}
Lee Xiong, Chenyan Xiong, Ye Li, Kwok-Fung Tang, Jialin Liu, Paul Bennett, Junaid Ahmed, and Arnold Overwijk.
\newblock Approximate nearest neighbor negative contrastive learning for dense text retrieval.
\newblock \emph{arXiv preprint arXiv:2007.00808}, 2020{\natexlab{a}}.

\bibitem[Xiong et~al.(2020{\natexlab{b}})Xiong, Li, Iyer, Du, Lewis, Wang, Mehdad, Yih, Riedel, Kiela, et~al.]{xiong2020answering}
Wenhan Xiong, Xiang~Lorraine Li, Srini Iyer, Jingfei Du, Patrick Lewis, William~Yang Wang, Yashar Mehdad, Wen-tau Yih, Sebastian Riedel, Douwe Kiela, et~al.
\newblock Answering complex open-domain questions with multi-hop dense retrieval.
\newblock \emph{arXiv preprint arXiv:2009.12756}, 2020{\natexlab{b}}.

\bibitem[Yu et~al.(2021)Yu, Zhu, Fang, Yu, Wang, Xu, Ren, Yang, and Zeng]{yu2021kg}
Donghan Yu, Chenguang Zhu, Yuwei Fang, Wenhao Yu, Shuohang Wang, Yichong Xu, Xiang Ren, Yiming Yang, and Michael Zeng.
\newblock Kg-fid: Infusing knowledge graph in fusion-in-decoder for open-domain question answering.
\newblock \emph{arXiv preprint arXiv:2110.04330}, 2021.

\bibitem[Yu(2022)]{yu2022retrieval}
Wenhao Yu.
\newblock Retrieval-augmented generation across heterogeneous knowledge.
\newblock In \emph{Proceedings of the 2022 Conference of the North American Chapter of the Association for Computational Linguistics: Human Language Technologies: Student Research Workshop}, pages 52--58, 2022.

\bibitem[Yu et~al.(2023)Yu, Zhang, Liang, Jiang, and Sabharwal]{yu2023improving}
Wenhao Yu, Zhihan Zhang, Zhenwen Liang, Meng Jiang, and Ashish Sabharwal.
\newblock Improving language models via plug-and-play retrieval feedback.
\newblock \emph{arXiv preprint arXiv:2305.14002}, 2023.

\bibitem[Zhang et~al.(2023)Zhang, Gong, Zhang, Li, Qiao, Ouyang, and Yue]{zhang2023meta}
Yiyuan Zhang, Kaixiong Gong, Kaipeng Zhang, Hongsheng Li, Yu Qiao, Wanli Ouyang, and Xiangyu Yue.
\newblock Meta-transformer: A unified framework for multimodal learning.
\newblock \emph{arXiv preprint arXiv:2307.10802}, 2023.

\bibitem[Zheng et~al.(2024)Zheng, Gou, Kil, Sun, and Su]{zheng2024gpt}
Boyuan Zheng, Boyu Gou, Jihyung Kil, Huan Sun, and Yu Su.
\newblock Gpt-4v (ision) is a generalist web agent, if grounded.
\newblock \emph{arXiv preprint arXiv:2401.01614}, 2024.

\bibitem[Zhu et~al.(2023)Zhu, Chen, Shen, Li, and Elhoseiny]{zhu2023minigpt}
Deyao Zhu, Jun Chen, Xiaoqian Shen, Xiang Li, and Mohamed Elhoseiny.
\newblock Minigpt-4: Enhancing vision-language understanding with advanced large language models.
\newblock \emph{arXiv preprint arXiv:2304.10592}, 2023.

\end{thebibliography}
}


\end{document}